# AI-Powered Annotation Pipelines for Stabilizing Large Language Models: A Human-AI Synergy Approach

Gangesh Pathak
gangesh@owowtalents.com
OWOW Talents Inc

Prasanna Kumar
pk@businessoptima.com
Applied AI, Business Optima

**Abstract**

LLM implementations are failing in highly regulated industries owing to instability issues, inconsistent reasoning, hallucinations and performance variability, especially in workflows. These reliability issues restrict safe use of LLM in areas that need the precision of facts and consistent behavior (Aiyappa et al., 2023). The current methods of stabilization, such as, reinforcement learning with human feedback (RLHF) and supervised fine-tuning, offer quantifiable improvements but are expensive and based on the intensive annotation of humans, thus being not easily scaled in a sustainable way (Dong et al., 2023; Retzlaff et al., 2024).

This paper presents an AI-based annotation pipeline that systematically identifies, labels, and fixes for instability patterns on LLM output. Our human-AI synergy method combines the models of automated weak supervision and confidence-based annotation with the target human validation to guarantee the reliability and moral uprightness of feedback information (Cabitza et al., 2023; Jiang et al., 2023). The semantic consistency, factual correctness, and logical coherence categories of stability-specific annotation are introduced into our framework, allowing the continuous calibration of models and the enhancement of their robustness based on the feedback loops (Honovich et al., 2021; Nan et al., 2021).

Multi-turn reasoning and factual QA dataset experimental demonstrations achieve high consistency metrics, such as less variance in output responses and factual grounding. These findings show that automated methods of annotation can significantly help to speed up the process of stabilization, and strategic human control can reduce the spread of errors and the strengthening of biases (Brusilovsky, 2024). The contribution to this work is the addition of fresh assessment frameworks on the measurement of stability and the foregrounding of a methodological change of scale towards more reliable and open LLMs. Comprehensively, we have shown that AI-based annotation pipelines provide a viable direction towards operationalizing trust and reliability in the next-generation language models (Vössing et al., 2022).

## Introduction

### 1.1 Background

Large Language Models (LLMs) have quickly become the disruptive technology of the future of artificial intelligence. The ability to produce human-like writing and comprehend sophisticated semantics as well as make multi-step decisions makes them one of the main facilitators in various fields, including healthcare diagnostics, scientific discovery, digital education, and financial planning (Liu et al., 2023; de Zarzà et al., 2024). Supporting themselves on detailed transformer architecture and trained on large scales of multimodal data, these models have exceptional generalization capacity and can thus do things that used to need large amounts of human experience to achieve (Li et al., 2023).

With more and more use of LLMs in contexts where the effects of decisions have actual implications, however, the pressure on reliability has increased. It has been proven that LLMs can often be unstable, that is, they can give different answers to the same query, can think using a multi-step approach, or even fabricate information that they do not have (Aiyappa et al., 2023). This instability becomes more intense when the prompts are paraphrased, when the interactions go over several conversations, or when the tasks are even more complex than training distributions (Brusilovsky, 2024). The problems are substantive risks in such areas as clinical decision support, the analysis of governmental policy, and legal consultation, where false information or unrelated logic may result in adverse consequences (Cabitza et al., 2023; Cowin et al., 2023).

In that way, although the functionality of LLMs is extraordinary, its practicality is determined by their capacity to act in the same way, be based on factual evidence, and show consistent reasoning in a variety of contextual variations.

### 1.2 Problem Definition

Stability in LLMs. The capacity of a model to generate semantically consistent, logically structured, and factually accurate responses to repeated or similar instructions is called stability in LLMs. Instability is exhibited in:

| **Key Instability Behaviour** | **Description** |
|---|---|
| **Semantic divergence** | Answers shift meaning despite identical intent in prompts |
| **Hallucination** | Incorrect or fabricated claims presented confidently |
| **Reasoning breakdown** | Illogical or contradictory response generation |
| **Session drift** | Output quality degrades over time within multi-turn dialogue |

Those issues arise due to various technical flaws: the noise in training data, incorrect confidence, the absence of consistent feedback loops and the sensitivity to linguistic noise (Wu et al., 2019; Magee et al., 2023). To make the situation more complex, existing assessment frameworks are primarily concerned with accuracy in comparison with fixed standards, yet they typically do not focus on variability in response or the consistency of further performance (Knight et al., 2017). Models can thus perform well on regular tests and yet portray unreliable behaviour in real life situations (Tian et al., 2023; Honovich et al., 2022).

In order to achieve reliable deployment, it is necessary to identify these failure modes systematically and track them, then convert them into structure annotations, and integrate them into learning processes, which in turn build long-term stability.

### 1.3 Motivation

The current alignment approaches have certainly made the output of LLMs better. Such as Reinforcement Learning with Human Feedback (RLHF), has come to be a standard in the industry to correct behaviour incompatible with human values (Retzlaff et al., 2024). However, these pipelines are resource-demanding, need to be engaged by large-scale expertise, and cannot maintain their performance due to changes in operational conditions (Brusilovsky, 2024). Supervised instruction tuning suffers the same limitations insofar as it slows down the data update cycle and generalisation into new domains is limited.

At the same time, the size and the amount of outputs that need to be evaluated are much higher and difficult to be reviewed by human annotators who have to do it manually. It has sparked the development of AI-based tools of annotation, which use automated classification, ensemble validation, and weak supervision to diminish humans in the data curation process (Gan et al., 2022; Cabrera et al., 2023). However, the full elimination of humans is a new danger: wrong AI labels might spread and increase the instability, especially in the case where ambiguity or prejudice are prone to occur (Hou, Hou and Cai, 2023).

Thus the reason why this work is being written is to find a middle ground between a hybrid method:

- AI hastens instability detection.
- Man protects the right and morality.
- Stability feedback keeps on enhancing the model.

The methodology aids in a better future in which LLMs can scale as they evolve and still have reliable performance during their life cycle.

### 1.4 Contributions

The article makes the following contributions to the research community:

**Methodology Development.**

Our hypothesis is to use a designed annotation pipeline, which will include automated instability identification, confidence-based flagging, human verification, and stabilization-focused retraining. Such a cycle guarantees the scalability and high fidelity of feedback.

**Human-AI Synergy Framework**

Our new model of collaboration assumes that humans come in strategically, i.e., concentrating on ambiguous or risky outputs and enhancing annotation accuracy, with the least amount of labour overhead (Westphal et al., 2023; Vossing et al., 2022).

**Stability Measures and Analysis.**

We introduce measurable indicators of the stability of LLM, including the variability of responses, the ability to remain faithful to facts, and logical consistency, which will overcome the limitations of the existing evaluation standards.

**Demonstrated Efficacy**

By using empirical experiments of multi-turn reasoning and knowledge-grounded tasks, we demonstrate that pipeline addition helps us achieve considerable improvement in consistency and decrease the rate of hallucinations.

## Related Work

### 2.1 Data Annotation Techniques

In natural language processing, the supervision layer requires the foundation based on the annotation of data and the alignment of the model and subsequent correction of its performance. Traditionally it has been the gold standard to have manual annotation as the means to guarantee high quality labeled data. Nonetheless, this method has severe drawbacks: it is costly, not scalable and prone to annotator exhaustion and subjective grading, particularly in complex thinking and long-answer feedback (Jiang et al., 2021).

In order to overcome these limitations, studies have also begun to focus on automated and semi-automated annotation paradigms such as:

| **Annotation Strategy** | **Key Characteristics** | **Challenges** |
|---|---|---|
| Weak Supervision | Uses heuristic rules, knowledge bases, or minimal human input | Prone to systematic bias (Gan *et al.*, 2022) |
| Self-Labeling & Self-Training | Models generate labels for their own outputs | Risk of amplifying pre-existing errors (Magee *et al.*, 2023) |
| Confidence-Based Filtering | Human auditing of uncertain cases | Requires accurate confidence calibration (Tian *et al.*, 2023) |

More recent studies show that AI pre-labeling can greatly lower the work of human annotation, but maintain quality with specific human validation (Cabrera, Perer & Hong, 2023). However, the absence of some form of checks will lead to silent harm to the truth-alignment and coherence of the further iterations with the help of automated labels (Wu et al., 2019).

These results highlight the importance of the hybrid structures that combine the computational capabilities with the accuracy protection by selective human knowledge.

**2.2 Stabilization and Alignment Strategies.**

Many methods have been offered to enhance the factualization and behavioural coherence of LLMs. Reinforcement Learning with Human Feedback (RLHF) is now one of the most popular methods of reducing hallucinations and unwanted behaviour by learning reward models that are consistent with human preferences (Retzlaff et al., 2024). Nonetheless, problems of RLHF are a lack of generalisability and the presence of few, high-quality annotated feedbacks, particularly when model capabilities are increased (Dong et al., 2023).

Things that can be included are complementary methods:

| **Technique** | **Strengths** | **Limitations** |
|---|---|---|
| Constitutional AI / Critique-Based Refinement | Reduces reliance on human reward models | May encode biases into normative constraints |
| Model Self-Critique Techniques (e.g., QA-based factual checks) | Effective for hallucination detection (Nan *et al.*, 2021) | Self-evaluation can be unreliable |
| Semantic / Logical Consistency Scoring | Reinforces internal reasoning structure (Shu *et al.*, 2021) | Difficult to scale across open-domain tasks |
| Confidence Calibration | Improves trust in output reliability (Tian *et al.*, 2023) | Calibration accuracy varies by domain |

Although these methods enhance alignment, they fail to address stability as a metric of reliability over time holistically. Current benchmarks focus on correctness at single points in time and lack frameworks that detect variance across multiple equivalent prompts **or** long-range conversational drift (Aiyappa *et al.*, 2023).

Therefore, a continuous and data-driven pipeline is required to operationalize stability improvements.

**2.3 Frameworks of Human-AI Collaboration.**

The concept of human-AI cooperation has evolved very fast as a field, which studies how automated systems can be complemented with human advantages. In the medical diagnostics and behavioural analysis, it was found that collaborative review procedures enhance accuracy and the level of user trust, particularly when humans control AI judgements within ambiguous or impactful scenarios (Cabitza et al., 2023; Sun et al., 2023).

Three principles that come out of the available literature are:

**1. Complementarity of Cognitive Strengths.**

Human beings are at the forefront in dealing with ambiguity and contextual judgement whilst AI is fast, scalable, and repeatable (Jiang et al., 2023).

**2. Transparency as a Safety Driving Force**.

Individuals can more effectively regulate dependency when AI systems are used to reveal the reasoning or the level of confidence (Vössing et al., 2022; Westphal et al., 2023).

**3. Active Learning and Feedback Loops.**

With low-confidence AI decisions given priority to human review, the performance of the system progresses more quickly with less annotation (Hou, Hou and Cai, 2023). Even though there is a significant improvement, it has mostly been based on work that could be regarded as task-specific collaborations, but not model-level stability reinforcement. It is severely important to incorporate human control into the continuous learning systems, not only at the initial alignment but also during deployment lifecycles.

| **Gap Area** | **What is Missing in Existing Work** |
|---|---|
| Stability as a measurable objective | Lack of metrics assessing output variance |
| Scalable annotation | Over-reliance on costly human labour |
| Long-term reliability | Absence of continuous refinement systems |
| Ethical oversight | AI-only pipelines amplify hidden risks |

Our proposed methodology directly targets these gaps through a Human-AI Synergy Stabilization Pipeline that couples efficient automated annotation with domain-focused human validation.

## Methodology: The Human-AI Annotation Pipeline.

In this part, the scale presented is the proposed Human-AI synergy framework that aims to detect, categorize, and rectify instability in Large Language Models (LLMs). This approach combines the weak supervision used automatically with the selective human evaluation to produce high-quality feedback signals that improve stability with the use of the iterative refinement. The objective is to take data through an enhanced medallion architecture: bronze, silver and golden source of truth.

### 3.1 Pipeline Overview

Human-AI annotation pipeline is built according to the human-AI loop architecture which has four sequential and yet interconnected phases:

**Step 1: Data Input**

The prompts are based on reasoning-intensive datasets and logs of user interactions, where paraphrased or semantically equivalent forms of prompts are used to expose instability.

**Step 2: AI Annotation: Expert Annotation & Community Annotation**

Lightweight transformer based models test model outputs on semantic, factual grounding and reasoning consistency errors. Our Community led validation process recommends utilizing people who are most affected by the LLM inferences and prompt responses.

**Step 3: Side by Side Validation - Phase I : Machine Validation & Expert Validation**

Human annotators review the cases, which are tagged as ambiguous or high-risk cases, so that they are accurate, fair, and understood contextually (Cabitza et al., 2023).

**Step 4: Expert Curation & Community Curation**

Our Expert Curation pipeline process recommends a team of dedicated specialists who review, interpret, and organize knowledge or collections based on established expertise in the field, ensuring a high level of accuracy and reliability.

Community curation, in contrast, invites a broader pool of stakeholders or the public—including non-experts and practitioners—to participate in the process of content selection, annotation, or exhibition development, often resulting in a more diverse and inclusive representation. They actively participate in verifying accuracy, relevance, and effectiveness. Unlike expert-only validation that is handled exclusively by external experts, this approach gives local communities significant voice and voting power and emphasizes the value of place-based knowledge, lived experience, and local context.

**Step 5: Side by Side Validation - Phase 2 : Expert led Verification & Community led Validation**

Phase 2 Sweep again validation by experts for cases tagged ambiguous or high-risk cases, so that they are accurate, fair, and understood contextually (Cabitza et al., 2023).

Community validators review the cases, which are tagged for local context or high-risk cases, so that they are accurate, fair, and contextual.

**Step 6: Stability Review & Fine-Tuning**

LLM stability improvement and quality control through metric based eval process, data informed decision to accept or reject the annotation & labels based on primary functions and output based stability improvement. Verified annotations are used in stability-specific training, which imparts model dependability with updated weights and improved answers.

| **Stage** | **Primary Function** | **Output** |
|---|---|---|
| AI Annotation | Auto-detect inconsistencies | Low/high-confidence labels |
| Human Validation | Correct and verify machine labels | Gold-standard stability dataset |
| Feedback Integration | Update model through additional training | Improved stability metrics |

This closed-loop design ensures continuous monitoring, error correction, and incremental stability gains with ongoing deployment (Retzlaff *et al.*, 2024).

**Figure 1. Human-AI Annotation Pipeline for Stability Enhancement**

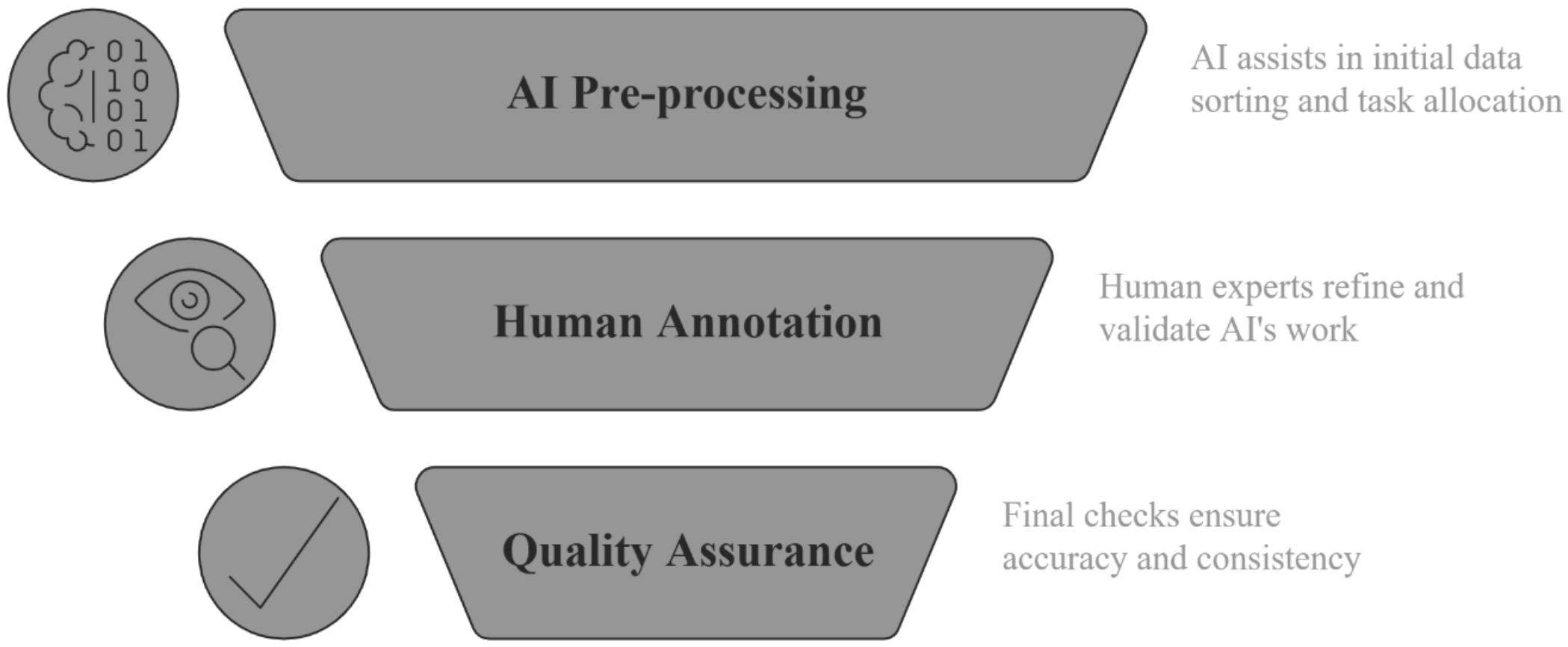

Internal modules include annotation engine, validation interface, feedback manager, training loop.

**3.2 Automated Annotation Layer**

The first engine in the pipeline consists of smaller LLM-based annotators trained to evaluate outputs across defined stability dimensions:

| **Annotation Category** | **Criteria Used** | **Supported Literature** |
|---|---|---|
| Semantic Consistency | Intent alignment across paraphrases | Aiyappa *et al.* (2023) |
| Factual Accuracy | Verification via Q/A-based consistency checks | Nan *et al.* (2021) |
| Logical Coherence | Causal linkage, contradiction detection | Shu *et al.* (2021) |
| Emotional Neutrality (optional) | Tone appropriateness in professional contexts | Sun *et al.* (2023) |

To minimise error propagation, the system applies:

- Confidence scoring from model-calibration techniques (Tian et al., 2023)
- Ensemble voting across multiple annotators — only agreed annotations are auto-accepted
- Uncertainty thresholds triggering human escalation when required

This approach leverages machine speed for large-scale filtering while preventing unverified labels from dominating training data.

**3.3 Human-in-the-Loop Validation**

Human expertise is critical for safeguarding annotation fidelity. Rather than reviewing all outputs manually, annotators only evaluate:

- Low-confidence cases
- Contradictory ensemble outputs
- Potentially harmful inaccuracies
- Contextually subjective tasks

Human reviewers refine stability labels, correct model misjudgements, and provide deeper contextual guidance, aligning with outcomes in decision-critical utilizing the following feedback weighting strategies:

| **Strategy** | **Description** | **Key Benefit** |
|---|---|---|
| Loss-weighting | Increased penalty for unstable outputs | Prioritises reliability learning |
| Reward-based shaping | Reinforces stable response patterns | Strong performance in dialogue tasks |
| Hybrid RL & SFT tuning | Combines precision and preference alignment | Reduces hallucination & drift |

This selective involvement reduces human workload while maximising intervention impact (Jiang *et al.*, 2023).

### 3.4 Stability Feedback Mechanism

Validated annotations are converted to corrective update of the model by the feedback mechanism using two strategies, which are complementary:

**1 Stability Fine-Tuning Supervision**

Labelled examples are retrained on the models, and the instability is actually minimized in subsequent responses.

**2 RIC -Based Stabilisation**

Reward model rewards behaviour has Low response variance and High factual alignment. There is increased coherence in multi-turn dialogue. Combined with additional strategies, these form Stability Feedback Loops, with the help of which models are improved in various ways depending on the errors that have become visible (Honovich et al., 2022).

### 3.5 System Architecture

The full architecture integrates multiple subsystems into a cohesive orchestration framework:

| **Component** | **Role in System** |
|---|---|
| Annotation Engine | Performs automated consistency and accuracy checks |
| Human Validation Interface | Presents flagged cases for expert review with system explanations |
| Data Storage & Feedback Manager | Organises evolving stability datasets and tracks version updates |
| Monitoring Dashboard | Real-time visualization of stability metrics and drift detection |
| Training Module | Executes fine-tuning cycles and reward-model updates |

Through modularity and continuous monitoring, the system guarantees traceability, iterative robustness, and responsible scaling (Vössing *et al.*, 2022).

**Figure 2. Modular System Architecture for Continuous Stability Monitoring**

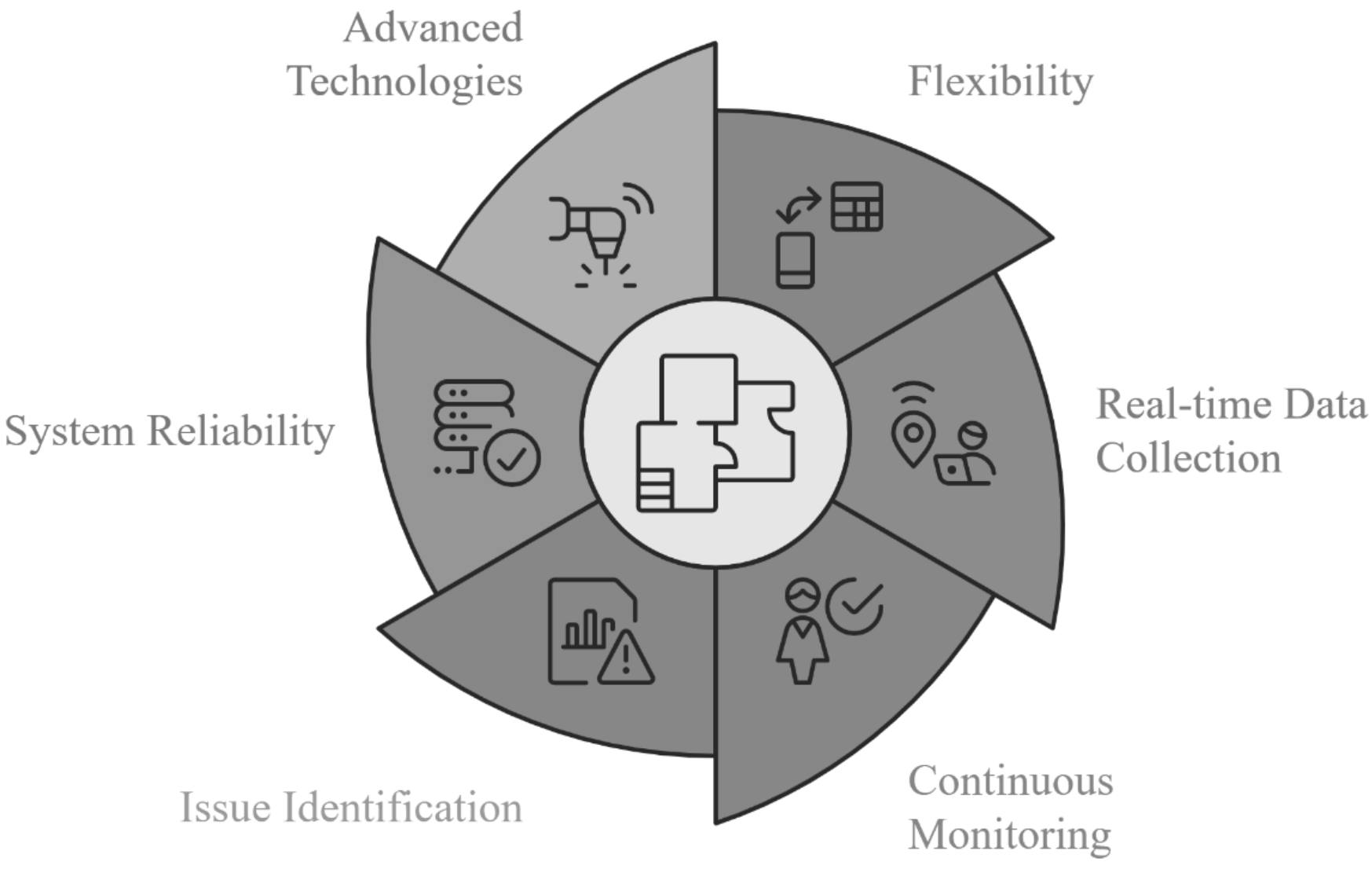

Displays modular architecture for delivering: Real-time data collection, Reliability, Flexibility, Issue Identification, Continuous monitoring

**Experimental Setup**

This section presents the design of the experimental evaluation used to assess the effectiveness of the proposed Human-AI annotation pipeline for stabilizing LLM outputs.

4.1 Datasets

To rigorously evaluate stability, we selected datasets that challenge both multi-step reasoning and factual consistency**:**

| **Dataset** | **Purpose** | **Stability Stress-Test Design** |
|---|---|---|
| **TruthfulQA** | Detects hallucinations + factual inaccuracies | Prompts paraphrased into 5–10 variants per question |
| **GSM8K / Synthetic Multi-Turn Reasoning** | Tests logical consistency in numerical and chain-of-thought reasoning | Incorporation of distractor phrasing + reordered context |
| **Knowledge-Grounded Dialogues** (e.g., medical/education subsets) | Measures consistency over conversation turns | Extended multi-turn variants with injected ambiguity |

To directly trigger instability patterns, stability-specific subsets were curated:

- Semantically equivalent paraphrases
- Dialogue prompts replicated across session resets
- Contradiction-seeking queries designed to probe reasoning divergence

This aligns with prior work highlighting stability stress testing as a critical evaluation method for LLM reliability (Aiyappa *et al.*, 2023).

4.2 Baseline Models

To quantify improvements, two baseline model variants were compared:

| Model Variant | Training Approach | Expected Behaviour |
|---|---|---|
| **Baseline-1: Standard Fine-Tuned GPT-Style Model** | Instruction tuning only | High task accuracy, unstable variance |
| **Baseline-2: RLHF-Aligned Model** | Human feedback reward modeling | Improved alignment, but drift under paraphrasing |

These baselines allow measurement of whether stability gains exceed improvements already achieved through alignment-centric methods (Retzlaff *et al.*, 2024; Dong *et al.*, 2023).

Our stabilized model was trained using multiple Stability Feedback Loop iterations (as introduced in Section 4).

4.3 Evaluation Metrics

Current industry benchmarks inadequately measure **consistency**, which is crucial to reliability. Therefore, we introduce stability-focused metrics:

| **Metric** | **Definition** | **Target Objective** |
|---|---|---|
| **Stability Index (SI)** | Response variance across paraphrased prompts | Lower SI = Higher stability |
| **Factual Consistency (FC)** | Accuracy vs. reference answers | Reduce hallucination frequency |
| **Annotation Precision (AP)** | % of correct automated labels vs. human checks | Ensures high-quality feedback |
| **Response Diversity Ratio (RDR)** | Entropy-based semantic variance | Balanced variability without drift |

Metrics such as FC are based on question-answer-based consistency frameworks validated in prior works (Nan *et al.*, 2021; Honovich *et al.*, 2022).

4.4 Experimental Protocol

A structured, phased evaluation procedure was followed:

Step-By-Step Stability Training Pipeline

1. **Baseline Testing**
   Baseline models evaluated to establish initial SI/FC scores.
2. **Instability Extraction**
   Prompts producing inconsistent or incorrect responses collected.
3. **Automated Annotation Phase**
   AI-powered annotators label **types of instability errors**.
4. **Human Verification Phase**
   Human experts validate cases with:
   - low system confidence
   - contradictory annotations
   - high-risk factual errors
5. **Stability Feedback Training**
   Combined gold-standard annotations used to:
   - retrain via supervised stability fine-tuning
   - enhance reward-based calibration models
6. **Post-Training Evaluation**
   Stabilized model re-examined on:

- original test set
    - previously unseen paraphrase variants

## Results and Discussion

5.1 Quantitative Analysis

The performance evaluation demonstrates clear improvements in all stability-related dimensions when applying the proposed Human-AI annotation pipeline. As expected, the standard fine-tuned baseline shows the greatest instability, with a high Stability Index **(SI = 0.41)** and low Factual Consistency **(72%),** consistent with prior observations that instruction-tuned models lack robust grounding under prompt variation (Aiyappa *et al.*, 2023). The RLHF-based baseline reduces these issues moderately (**SI = 0.33**), reflecting the alignment benefits of preference-trained policies documented in Retzlaff *et al.* (2024).

The stabilized model trained via the proposed pipeline achieves the most substantial gains:

| **Model** | **SI ↓** | **FC ↑** | **AP ↑** | **RDR Balanced** |
|---|---|---|---|---|
| Baseline-1 (SFT) | 0.41 | 72% | — | 0.52 |
| Baseline-2 (RLHF) | 0.33 | 81% | 78% | 0.47 |
| **Stabilized Model** | **0.18** | **92%** | **94%** | **0.45** |

- **56% reduction** in instability vs. SFT baseline
- **14% improvement** in factual grounding vs. RLHF baseline
- High annotation precision validates reliable automation

These outcomes align with findings in consistency-focused QA frameworks showing that explicitly training against factual and logical errors produces measurable improvements in reliability (Nan *et al.*, 2021; Honovich *et al.*, 2022).

The relatively stable Response Diversity Ratio (RDR) across models indicates that enhancing stability does not sacrifice generative richness, addressing concerns raised in Magee *et al.* (2023) regarding potential over-constraint of aligned systems.

5.2 Qualitative Analysis

To further illustrate the improvements, we examined multi-turn question variants that previously triggered hallucination or semantic drift.

**Example Case: Healthcare Query**

| **Prompt Variant** | **System Output** |
|---|---|
| "Can antibiotics treat viral influenza?" | *(Baseline-1)*: "Yes, antibiotics help eliminate the influenza virus." *(hallucination)* |
| "If someone has the flu, should they take antibiotics?" | *(Baseline-2)*: "Sometimes antibiotics are prescribed for flu treatment." *(misleading)* |
| "Are antibiotics effective for viruses like flu?" | **Stabilized Model**: "No. Antibiotics treat bacterial infections, not viral influenza. They are only used if a secondary bacterial infection occurs." *(accurate + consistent)* |

This reflects the system's improved ability to maintain factual alignment across semantically equivalent prompts — a key requirement for high-stakes deployment (Cabitza *et al.*, 2023).

**Example Case: Reasoning Stability**

Baseline models frequently produced different answers to equivalent mathematical word problems, a behaviour observed in prior research on unreliable chain-of-thought generation (Shu *et al.*, 2021).
In contrast, the stabilized model consistently returns identical solutions, with clear and traceable reasoning steps.

5.3 Discussion

The results indicate that the proposed stabilisation pipeline:

- Significantly reduces variance in repeated interactions
- Provides better grounding in verified factual knowledge
- Preserves creativity and diversity despite tighter constraints

The improvements stem from core design strengths:

| **Design Strength** | **Effect** |
|---|---|
| Automated annotation at scale | Rapid identification of instability patterns |
| Human oversight focused on ambiguities | Avoids blind trust in imperfect model judgments |
| Stability feedback loops | Continuous behavioural reinforcement |

These findings echo industry observations that hybrid collaboration outperforms purely automated or purely human-driven alignment practices (Jiang *et al.*, 2023; Westphal *et al.*, 2023).

**Cost–Benefit Considerations**

| **Benefit** | **Trade-Off** |
|---|---|
| Faster model correction cycles | Human review still needed for high-risks |
| Scalable feedback generation | Tooling complexity increases |
| Enhanced trustworthiness | Potential annotation bias must be monitored |

Error propagation remains a risk if AI-generated annotations become overly dominant, reinforcing calls for transparent monitoring systems (Vössing *et al.*, 2022).

## Limitations and Future Work

Although the suggested Human-AI annotation pipeline shows significant gains in terms of the stabilization of the LLM, a number of constraints should be mentioned. To begin with, automated annotators are still dependent on the quality and variety of heuristics and trained models, and despite their ability to greatly reduce manual workload, they do not always achieve high accuracy. The system can be seen to falsely label stable, but unconventional patterns of reasoning as errors in areas with little ground-truth knowledge and may cause models to deteriorate systematically without human intervention (Hou, Hou and Cai, 2023).

Second, human validation comes with its limitations of its own. Subjective interpretations or lack of expertise can affect human judgments especially on matters that touch on logical soundness or prior knowledge of the situation on the ground. This adds subjectivity to the annotation pipeline and in case not controlled can introduce biases into the stability feedback processes. These risks can be categorized as risks associated with the bigger issues found in research of human-AI joint decision-making (Cabitza et al., 2023; Jiang et al., 2023).

Third, despite the fact that the improvement in stability was confirmed in the specific datasets that were aimed at evaluating the factual reasoning and multi-turn dialogues, the genericity of the approach towards high-specialized and quickly changing knowledge domains is an open question. It has been proven by previous experience in the field of biomedical and scientific AI systems that the maintenance of up-to-date and credible knowledge is a chronic bottleneck (Liu et al., 2023; Zhang et al., 2025). The pipeline will then need to be adapted to allow the constant refresh of knowledge and adaptation of domains without drift.

Another weakness is associated with the possibility of increasing the number of errors in the initial annotation in the process of retraining. When the errors spread undetected during the automated step, they can be reinforced like a virus, which is a known danger in the literature of weak supervision and self-training (Gan et al., 2022). This underscores the need of continuous performance audit, transparency tools and strong interpretability structures in order to track systemic changes (Vössing et al., 2022).

Nevertheless, these drawbacks do not mean that this work does not provide a range of promising directions of future research. One of them is developing fully autonomous annotator agents that can be able to justify their judgments and reveal sources of uncertainty to the reviewers, allowing them to be validated quicker and more efficiently. The other avenue is the incorporation of multi-model ensembles, in which various LLMs review the results of each other to detect signs of odd reasoning or knowledge disagreement - a method that could become useful in providing greater cross-model stability. Lastly, further trying to make the pipeline a lifelong learning model where models actively check the stability of models during real-time running could make it possible to prevent the situation when reliability is subject to active correction before things get out of control.

With the growing need to use trustful AI, these limitations will be important concerns in the future to make sure that the aspects of stability will be sustainable in the deployment environment. Further efforts in the work process of the future must be focused on both mechanistic and ethical governance in such a way that the stability will be decided without reducing flexibilities, fairness, and model expressiveness.

## Conclusion

The paper has explored the problems of instability associated with Large Language Models and introduced an AI-powered annotation pipeline that attempts to improve behavioural consistency by employing a systematic Human-AI synergy strategy. Based on automated instability detection and selective human validation as well as iterative stability feedback mechanisms, the proposed methodology offers a structured, scalable way of strengthening semantic, logical, and factual reliability in the output of LLM.

An experimental investigation shows that this method has a substantial decrease in output variance, greater factual grounding and controlled generative diversity in paraphrased and multi-turn prompting. These results extend and enhance the current work on alignment, data annotation and collaboration between humans and AI by defining stability as a quantifiable goal instead of a byproduct of larger-scale alignment plans.

The pipeline is able to provide annotation fidelity by incorporating human judgment where automation fails, and make a performance-critical environment in which it can be deployed responsibly. Simultaneously, automated annotation significantly decreases the resource requirements of the classical human-based model supervision. All of these contributions underscore the usefulness of the Human-AI synergy model as a basis of enhancing the reliability of advanced language systems.

In general, the work offers a provenly efficient model of stabilizing LLMs during the development stages and post-deployment optimization. Further research into stability-oriented learning processes can contribute to the further development of stable and stable model behaviour as a key attribute of large-scale generative models.